\documentclass{article}
\usepackage{graphicx} 
\usepackage{tabularx}
\usepackage{amsmath}
\usepackage{xspace}
\usepackage{booktabs}
\usepackage{tcolorbox}

\usepackage{colortbl}
\usepackage{graphicx}
\usepackage{enumitem}

\usepackage{wrapfig}

\usepackage{amsmath}
\usepackage{amsfonts}
\usepackage{bm}
\usepackage{vcell}

\usepackage[utf8]{inputenc} % allow utf-8 input
\usepackage[T1]{fontenc}    % use 8-bit T1 fonts
\usepackage{hyperref}       % hyperlinks
\usepackage{url}            % simple URL typesetting
\usepackage{booktabs}       % professional-quality tables
\usepackage{amsfonts}       % blackboard math symbols
\usepackage{nicefrac}       % compact symbols for 1/2, etc.
\usepackage{microtype}      % microtypography
\usepackage{xcolor}         % colors

\usepackage[preprint]{corl_2024} % Uncomment for pre-prints (e.g., arxiv); This is like ``final'', but will remove the CORL footnote.

\usepackage{booktabs}
\usepackage{graphicx}
\usepackage{subcaption}
\usepackage{tabularx}
\usepackage{graphicx}
\usepackage{amsmath}
\usepackage{amssymb}
\usepackage{corl_2024} % Uncomment for pre-prints (e.g., arxiv); This is like ``final'', but will remove the CORL footnote.

\newcommand{\methodLong}{{\textsc{Manipulate-Anything}\xspace}}
\newcommand{\lm}{{\textsc{VLM}\xspace}}

\title{ \includegraphics[height=6mm]{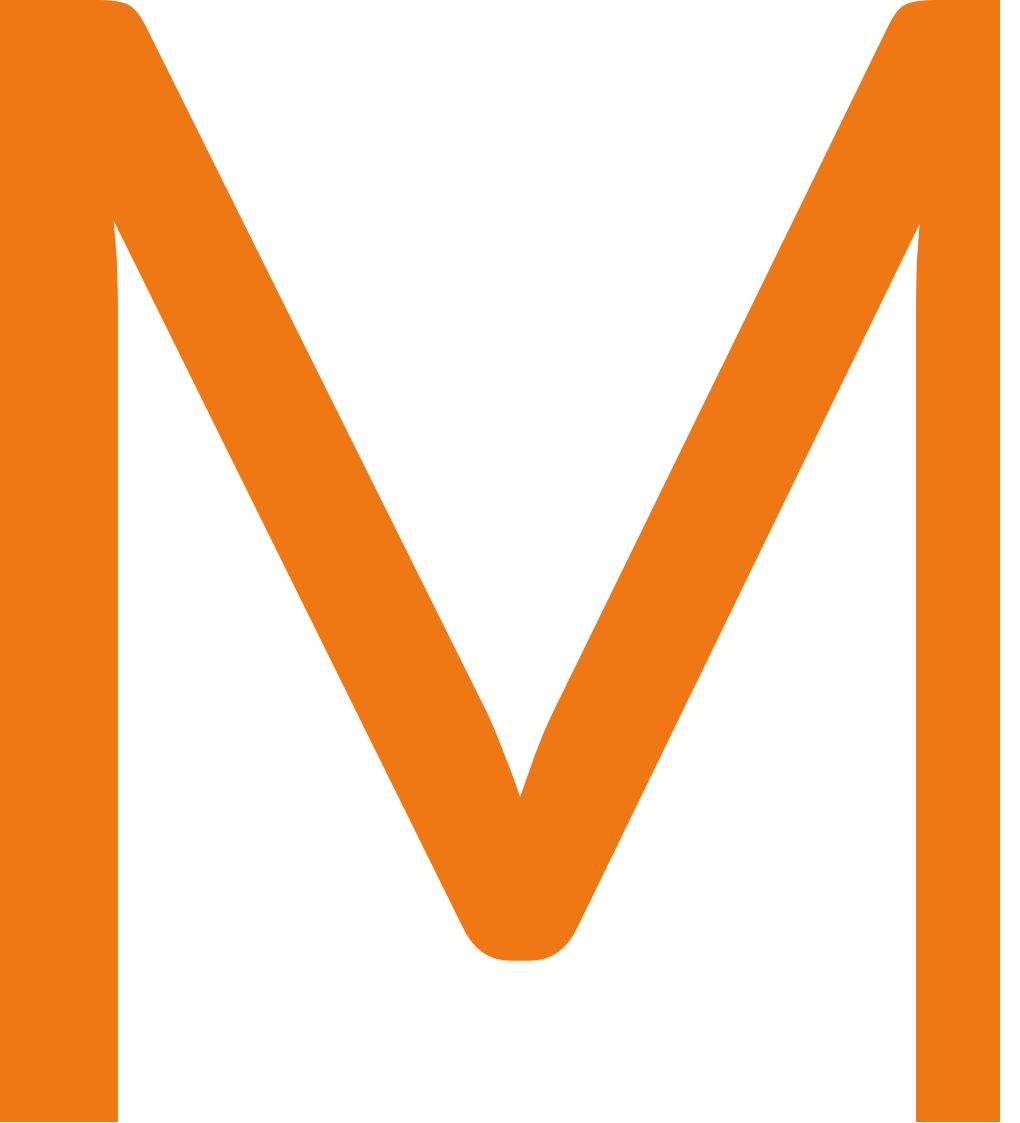}anipulate-\includegraphics[height=6mm]{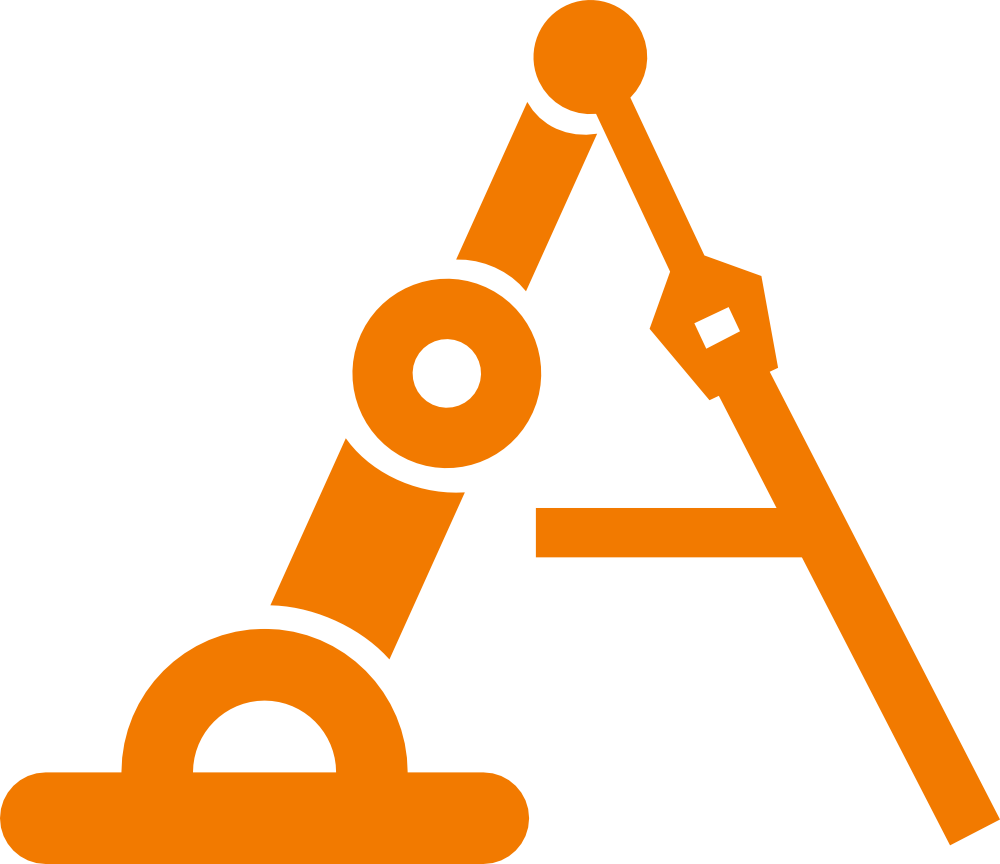}nything: Automating Real-World Robots using Vision-Language Models}

\author{ Jiafei Duan~$^{1}$\thanks{Equal contribution} \hspace{4px} Wentao Yuan~$^{1*}$ \hspace{4px} Wilbert Pumacay~$^2$ \hspace{4px} Yi Ru Wang~$^1$ \\
\textbf{Kiana Ehsani~$^3$} \hspace{4px} \textbf{Dieter Fox~$^{1, 4}$}\hspace{4px} \textbf{Ranjay Krishna~$^{1, 3}$}
\\
$^1$University of Washington \hspace{6px} $^2$Universidad Católica San Pablo \hspace{6px} \\
$^3$Allen Institute for Artificial Intelligence $^4$NVIDIA \hspace{6px}\\}

\begin{document}
\maketitle

\begin{abstract}
Large-scale endeavors like RT-1\cite{brohan2022rt} and widespread community efforts such as Open-X-Embodiment \cite{open_x_embodiment_rt_x_2023} have contributed to growing the scale of robot demonstration data. However, there is still an opportunity to improve the quality, quantity, and diversity of robot demonstration data. Although vision-language models have been shown to automatically generate demonstration data, their utility has been limited to environments with privileged state information, they require hand-designed skills, and are limited to interactions with few object instances.
We propose \methodLong, a scalable automated generation method for real-world robotic manipulation.
Unlike prior work, our method can operate in real-world environments without any privileged state information, hand-designed skills, and can manipulate any static object.
We evaluate our method using two setups.
First, \methodLong\ successfully generates trajectories for all $7$ real-world and $14$ simulation tasks, significantly outperforming existing methods like VoxPoser. 
Second, \methodLong's demonstrations can train more robust behavior cloning policies than training with human demonstrations, or from data generated by VoxPoser \cite{huang2023voxposer}, Scaling-up \cite{ha2023scaling} and Code-As-Policies \cite{liang2023code}.
We believe \methodLong\ can be a scalable method for both generating data for robotics and solving novel tasks in a zero-shot setting.  Project page: \href{https://robot-ma.github.io/}{robot-ma.github.io}.
\end{abstract}
\vspace{-1em}


\section{Introduction}
\vspace{-1em}

% THE DATA WE NEED
The success of modern machine learning systems fundamentally relies on the \textit{quantity}~\cite{kaplan2020scaling, cherti2023reproducible, hoffmann2022training, schuhmann2022laion, ramanujan2024connection, udandarao2024no}, \textit{quality}~\cite{gadre2024datacomp, zhou2024lima, nguyen2024improving, nguyen2022quality, lee2021deduplicating}, and \textit{diversity}~\cite{fang2022data, gururangan2020don, tian2023learning, xu2023demystifying, chen2023meditron} of the data they are trained on. The availability of large-scale internet data made possible significant advances in vision and language~\cite{deng2009imagenet,lin2014microsoft,schuhmann2022laionb}. However, the dearth of data has prevented similar advancements in robotics. Human demonstration collection methods do not scale to sufficient \textit{quantity} or \textit{diversity}.
Projects like RT-1~\cite{brohan2022rt} demonstrated the utility of high-\textit{quality} human data collected over 17 months. Others have developed low-cost hardware for data collection~\cite{chi2024universal,wang2024dexcap,duan2023ar2}. However, all these procedures require expensive human data collection. 
% \textcolor{red}{In an effort to diversify demonstration data, Open X-Embodiment project collected 1 million trajectories collected through a participatory effort by 34 research labs~\cite{open_x_embodiment_rt_x_2023}. Despite the wide-spread effort, the dataset only contains 20 tasks.}

% EXISTING MLM METHODS DON't SCALE
Automated data collection methods do not scale to sufficient \textit{diversity}.
With the advent of vision-language models (VLMs), the robotics community has been abuzz with new systems that leverage VLMs to guide robotic behavior~\cite{singh2023progprompt,liang2023code,wang2023gensim,wang2023genbot,huang2023voxposer,ha2023scaling,nasiriany2024pivot}. In these systems, VLMs decompose tasks into language plans~\cite{singh2023progprompt,liang2023code} or generate code to execute predefined skills~\cite{huang2024copa,huang2023voxposer}.
Though successful in simulation, these methods underperform in the real world~\cite{huang2024copa,huang2023voxposer}.
Some methods rely on privileged state information only available in simulation~\cite{wang2023gensim,ha2023scaling}, require hand-designed skills~\cite{wang2023genbot},
% Some methods rely on privileged state information, which is only available in simulation~\cite{wang2023gensim,ha2023scaling}; others require hand-designed predetermined skills that can be invoked by the LLM~\cite{wang2023genbot}.
or are also limited to manipulating a fixed set of object instances with known geometric shape~\cite{huang2023voxposer,huang2024copa}.

As VLMs improve in performance, and with the vast common-sense knowledge they have shown to possess, could we harvest their capabilities for diverse task completion and scalable data generation? The answer is yes -- with careful system design and the right set of input and output formulations, we can not only use VLMs as a means to successfully perform \textit{diverse} tasks in a zero-shot manner, but also generate \textit{quality} data at a high \textit{quantity} to train behaviour cloning policies.

\begin{figure}[t]
  \centering
  \vspace{-1em}
  \includegraphics[width=\linewidth]{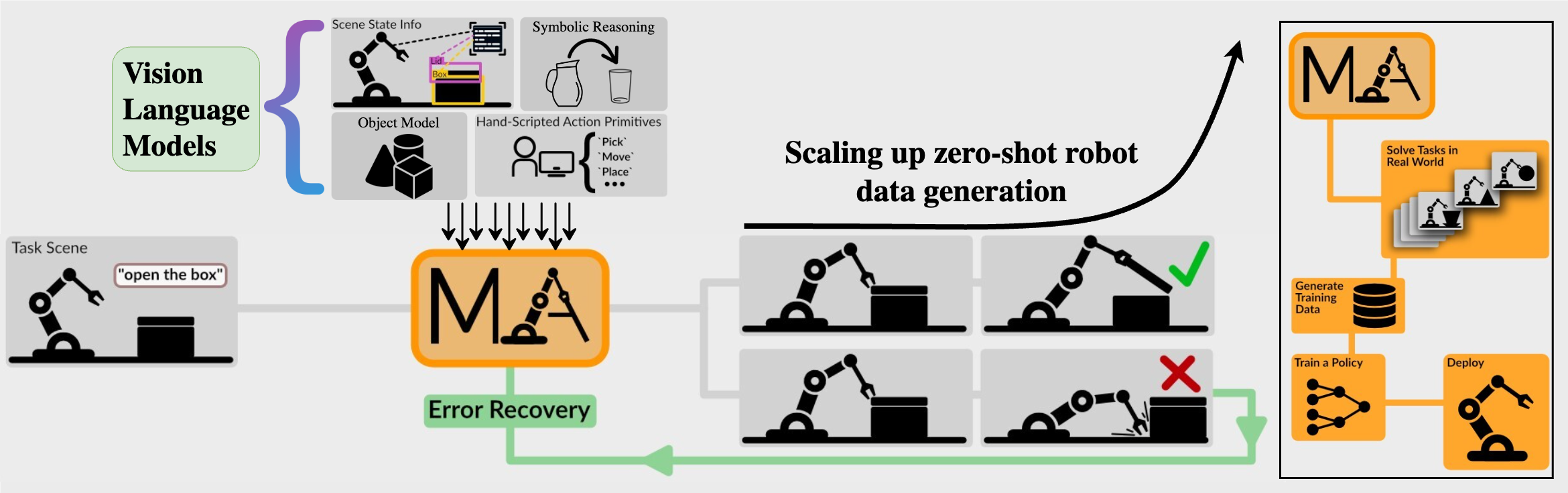}
  % Replace with your image file
  \caption{\methodLong\ is an automated method for robot manipulation in real world environments. Unlike prior methods, it does not require privileged state information, hand-designed skills, or limited to manipulating a fixed number of object instances. It can guide a robot to accomplish a diverse set of unseen tasks, manipulating diverse objects. Furthermore, the generated data enables training behavior cloning policies that outperform training with human demonstrations.}
  \vspace{-1em}

  \label{fig:2}
\end{figure}
% WE INTRODUCE A METHOD THAT DOES.
\textbf{We propose \methodLong\, a scalable automated demonstration generation method for real-world robotic manipulation.} 
\methodLong\ produces high \textit{quality} data, at large-\textit{quantities} (if needed), and can manipulate a \textit{diverse} set of objects to perform a \textit{diverse} set of tasks.
When placed in a real world environment and given a task (e.g., ``open the top drawer'' in Figure~\ref{fig:2}), \methodLong\ effectively leverages VLMs to guide a robotic arm to complete the task.
Unlike prior methods, it doesn't need privileged state information, hand-designed skills, or limited to specific object instances. Not relying on privileged information makes \methodLong\ environment-agnostic. \methodLong\ plans a sequence of sub-goals and generates actions to execute the sub-goals. It can verify whether the robot succeeded in the sub-goal using a verifier and re-plan from the current state if needed. This error recovery enables mistake identification, re-planning, and recovering from failure. It also injects recovery behavior into the collected demonstrations. We further enhanced the VLMs' capabilities by incorporating reasoning from multi-viewpoints, significantly improving performance.

% WHAT EXPERIMENTS WE RUN
We showcase the utility of \methodLong\ through two evaluation setups. First, we show that it can be prompted with a novel, never-before-seen task and complete it in a zero-shot manner. We quantitatively evaluate across $7$ real-world and $14$ RLBench~\cite{james2020rlbench} simulation tasks and demonstrate capabilities across many real-world everyday tasks (refer to supplementary). Our method significantly outperforms VoxPoser~\cite{huang2023voxposer} in 10/14 simulation tasks for zero-shot evaluation. It also generalizes to tasks where VoxPoser completely fails because of its limitation to specific object instances. Furthermore, we demonstrated that our approach can solve real-world manipulation tasks in a zero-shot manner, achieving a task-averaged success rate of 38.57\%. Second, we show that \methodLong\ can generate useful training data for a behavior cloning policy. We compare \methodLong\ generated data against ground truth hand-scripted demonstrations as well as against data from VoxPoser\cite{huang2023voxposer}, Scaling-up \cite{ha2023scaling} and Code-As-Policies~\cite{liang2023code}. Surprisingly, policies trained on our data outperforms even human hand-scripted data on 5 out of 12 tasks and performs on par for 4 more when evaluated with RVT-2. Meanwhile, the baselines are unable to generate the training data for some of tasks. \methodLong\ demonstrates the broad possibility of large-scale deployment of robots across unstructured real-world environments. It also highlights its utility as a training data generator, aiding in the crucial goal of scaling up robot demonstration data.
\vspace{-1em}

\section{Related work}
\vspace{-1em}

\methodLong\ enables scaling of robotic manipulation data using VLMs. As such, we review recent efforts in 1) scaling manipulation data, and 2) applications of VLMs in robotics. 

\noindent \textbf{Scaling manipulation data.} When deploying vision and language-based control policies for real-world applications, a significant challenge revolves around acquiring data. 
% This data encompasses trajectories which include inputs like vision, action labels (target actions and gripper commands), and task descriptions conveyed through language instructions. 
Traditionally, a convenient avenue to collect such trajectories is through human annotations for action (i.e. through teleoperation) and language labeling \cite{shridhar2023perceiver, shridhar2022cliport, brohan2023can}, however, this approach is limited to scale. To address this limitation and achieve autonomous scalability, prior works employ vision-language models or procedural generate language annotations in simulated environments \cite{ha2023scaling, ehsani2023imitating,ahn2024autort}. For action labels, strategies range from random exploration to learned policies \cite{wang2023robogen}. While human egocentric videos are relevant, they lack action labels and require cross-embodiment transfer \cite{grauman2022ego4d}. Another strategy involves model-based policies, such as task and motion planning (TAMP) \cite{garrett2021integrated}. Our approach extends these methods by incorporating common-sense knowledge from large language models (LLMs) and vision language models (VLMs), by providing a framework which combines the strengths of VLMs, object pose prediction, and dynamic retry to synthesize demonstrations in simulated and real environments.

\noindent \textbf{Language models for robotics.} In the field of robotics, large language models have found diverse applications, including policy learning \cite{xie2023text2reward,zhang2022sprint,wang2024rl}, task and motion planning \cite{lin2023text2motion, huang2022inner}, log summarization \cite{liu2023reflect}, policy program synthesis \cite{liang2023code}, and optimization program generation \cite{singh2023progprompt}. Previous research has also explored the physical grounding capabilities of these models \cite{huang2023voxposer,huang2024copa,zhang2023bootstrap}, while ongoing work investigates their integration with task and motion planners to create expert demonstrations \cite{ha2023scaling}. \cite{brohan2023can} attempted to collect extensive real-world interaction data, with short-horizon trajectories. \cite{liu2024moka} proposed a key-point based visual prompting method for real-world manipulation, through predicting affordances and corresponding motions. Our work complements the existing line of works, by leveraging the high-level planning capabilities of language models, scene understanding capabilities of vision language models, and action sampling, to enable synthesis of robot trajectories, which include language, vision, and robot state, given arbitrary tasks and environments.

\vspace{-1em}

\section{\methodLong}
\vspace{-1em}

\begin{figure}[t]
\vspace{-1em}
  \centering
  \includegraphics[width=\linewidth]{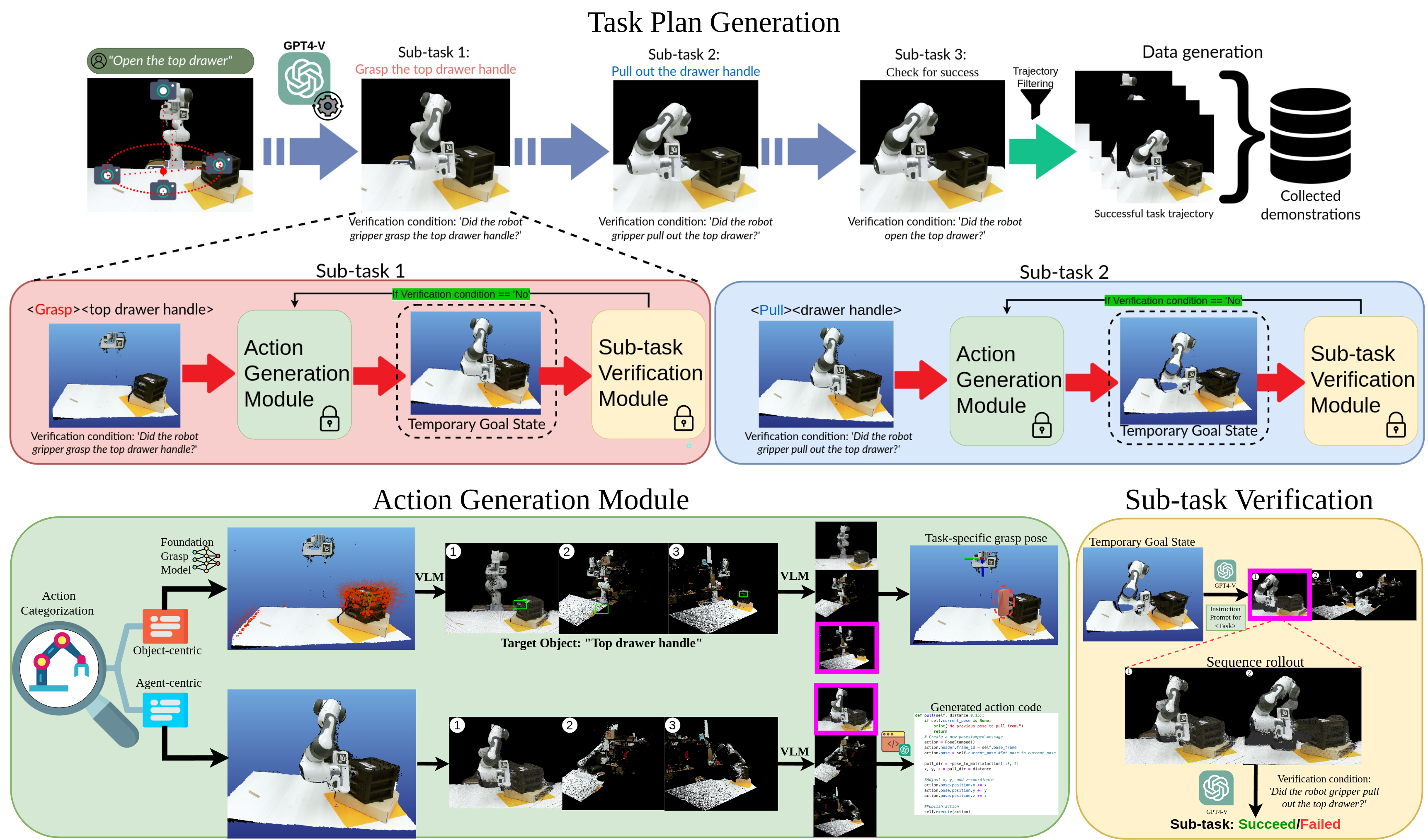} 
  % Replace with your image file
  \caption{\textbf{Manipulate Anything Framework.} The process begins by inputting a scene representation and a natural language task instruction into a VLM, which identifies objects and determines sub-tasks. For each sub-task, we provide multi-view images, verification conditions, and task goals to the action generation module, producing a task-specific grasp pose or action code. This leads to a temporary goal state, assessed by the sub-task verification module for error recovery. Once all sub-tasks are achieved, we filter the trajectories to obtain successful demonstrations for downstream policy training.}
  \label{fig:2}
  \vspace{-1em}

\end{figure}

\begin{figure}[t]
  \centering
  \includegraphics[width=\linewidth]{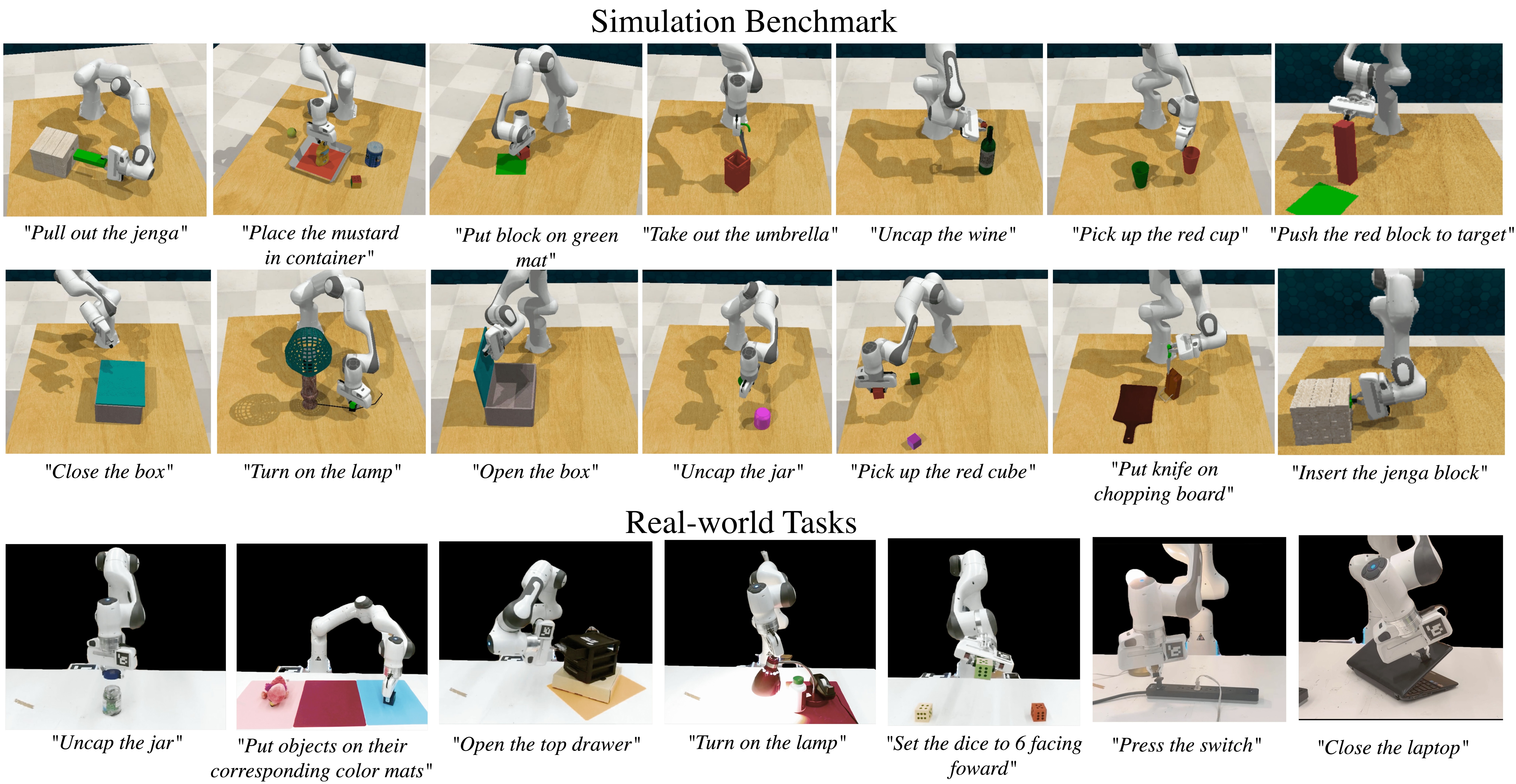}
  % Replace with your image file
  \caption{\methodLong\ is an open-vocabulary autonomous robot demonstration generation system. We show zero-shot demonstrations for 14 tasks in simulation and 7 tasks in the real world.}
  \label{fig:example}
  \vspace{-1em}

\end{figure}

We propose \methodLong, a framework that solves everyday manipulation tasks conditioned on language. Under the hood, \methodLong\ leverages VLMs to decompose tasks into sub-tasks, generates code for new skills or task-specific grasp pose, and verifies the success of each sub-task (Figure~\ref{fig:example}). Note that due to the modularity aspect of our framework, \methodLong\ will continue to improve as the underlying VLMs continue to improve.
% While using MLMs for robotic manipulation has recently been explored~\cite{huang2023voxposer,wang2023gensim,wang2023genbot}, our framework doesn't need \textit{any prior knowledge of the task, any predefined skills, or any privileged information}. 

% Our framework a) enables large-scale data generation, b) can work on a diverse set of tasks, and c) is capable of generating a diverse set of new skills.

% \kiana{We are emphasizing on this we should have experiments to back it up}
% operates in the real world.
% for robotic manipulation execution in the real-world . 
% Our approach  that have been pretrained on extensive internet data~\cite{li2022blip,zhu2023minigpt,bai2023qwen}. When utilized appropriately, VLMs exhibit robust vision capabilities for object grounding and spatial reasoning. 

% \vspace{-1em}
% \looseness-1
\subsection{Task plan generation}
\label{sec:task_plan}
\vspace{-1em}

\methodLong\ takes as input any task described by a free-form language instruction, $\mathbf{T}$ (e.g., \textit{`open the top drawer'}). 
Creating robot trajectories that adheres to $\mathbf{T}$ is challenging due to its potential complexity and ambiguity, requiring a nuanced understanding of the current environment state. 
Given $\mathbf{T}$, and an image of the scene, we apply a \lm\ to first identify task-relevant objects in the scene, appending them to a list. 
Subsequently, we use a LLM along with those information to decompose the main task $\mathbf{T}$ into a series of discrete, smaller sub-tasks, represented as $\mathbf{T}_i$, along with the corresponding verification conditions $v_i$, where $i$ ranges from 1 to $n$. 
For instance, the above task could be decompose into sub-tasks include \textit{`grasp the drawer handle'} or \textit{`pull open the drawer'}, and verification conditions are \textit{`did the robot grasp the handle?'} or \textit{`is the drawer opened?'}.
This transforms the instruction $\mathbf{T}$ into a sequence of specific sub-tasks $\{(\mathbf{T}_1, v_1), (\mathbf{T}_2, v_2), \dots, (\mathbf{T}_n, v_n)\}$. For each sub-task, \methodLong\ generates desired actions (\S~\ref{sec:action_generation}) and verifies them against the corresponding conditions to ensure successful completion of that sub-task(\S~\ref{sec:verifier}). This verification step also allows \methodLong\ to recover from mistakes and attempt again in the case of failure.

% Given a high-level task description, $\mathbf{I}$, e.g. `\emph{open the top drawer}', and a 3D scene in real or simulation \cite{james2020rlbench}, we apply a VLM to first identify and detect task-relevant objects in the scene. 
% These identified objects are input into a prompt template adapted from ProgPrompt \cite{singh2023progprompt}, which provides a structure for the available motion primitives and objects to interact with.  Subsequently, we leverage VLM to generate a task plan, $\{\mathbf{T}_1, \mathbf{T}_2, \dots, \mathbf{T}_n\}$, where each sub-task \( \mathbf{T}_i = [a,o, v]\), encompassing primitive actions, \( a \), target object, \( o \), and a verification condition, \( v \). Using this plan, an agent is initialized to perform the specified sub-tasks sequentially. \kiana{we need to rewrite this, this sentence can kill us, this primitive action contradicts our whole claim on no need to define primitve actions}
% \vspace{-1em}

\subsection{Multi-viewpoint VLM selection}
\label{sec:viewpoin}
\vspace{-1em}
Many prior works that investigated VLM's reasoning capability found that they do not work well given a single viewpoint \cite{liu2024coarse,yuan2024robopoint,hong20233d}. In robotic manipulation, we leverage multiple viewpoints from either more than one camera or re-rendering to minimize object occlusion \cite{goyal2023rvt,wang2023mvtrans}. Leveraging these insights, we proposed a multi-viewpoint selection phase via VLM before using the selected viewpoints for either action generation or sub-task verification for \methodLong\. We concatenate all available viewpoints from the current observations, either from multiple cameras or re-rendered viewpoints from a single RGB-D, into a single frame. Numbers are annotated on the top left of each concatenated frame to correspond to a specific viewpoint. Using the concatenated multi-viewpoints, we then query the VLM to choose an ideal viewpoint conditioned on the sub-task. Therefore, for both agent-centric and object-centric action generation, we render multiple viewpoints of the scene and query VLMs to choose an ideal viewpoint for either generating actions given the sub-task or verifying if the sub-task verification condition has been met as shown in Figure \ref{sec:task_plan}

\subsection{Action generation module}
\label{sec:action_generation}
\vspace{-1em}

Given a sub-task, the desired output from the action generation module is a sequence of low-level actions represented as a 6 DoF end-effector pose. 
The actions can be categorized into two sets: agent-centric or object-centric. Based on the generated sub-task, the LLM planner will classify it as either agent-centric or object-centric actions.
For agent-centric actions, it will modify the agent's state; e.g., it can move the robot's end-effector from the current state (e.g., ``rotate $90^{\circ}$''). 
We first employed our multi-viewpoint selection to sample out the most optimal viewpoints to provide the VLM with along with the in-context learning technique to generate the code for synthesizing the desired motion. Unlike prior methods that use only language models to generate code~\cite{liang2023code}, our approach utilizes VLM to understand and reason about object locations and the scene, which helps to ground the generation in the current state of the scene. This advantage is demonstrated in the ablation studies in the Appendix.

% A single view point might be insufficient to provide the \lm\ with enough information to perform the task (e.g., some views might be occluded by the robot arm). Therefore, we \textbf{render multiple viewpoints of the scene and query \lms\ to choose an ideal viewpoint} given the sub-task. For example, if the task is to open a drawer, the view in which the handle of the drawer is visible would be preferred. 
% After the best view point is chosen, the grasping poses can be filtered limited to the poses visible in that view point. The \lm\ chooses the optimal grasp pose and a simple motion planner can be used to move the robot to the desired grasping pose.

For object-centric actions, it is often used to generate a task-specific grasp pose for grasping a certain object (e.g., "grasp a knife for cutting") or to synthesize a pre-action pose for non-prehensile tasks by allowing the \lm\ to assign translation offsets to task-specific grasp poses. To obtain an object-centric action for a given sub-task, we first use an \textbf{object-agnostic grasp prediction model}~\cite{yuan2023m2t2} to generate all possible 6-DOF grasping poses in the scene. These poses are not conditioned on task specifications and may include invalid grasp poses for the given task. Next, we use the \lm\ to obtain the bounding box of the target object parts conditioned on the task specification from all available or re-rendered viewpoints (e.g., if the task is "grasp a knife," the \lm\ will detect the handle of the knife and generate a bounding box for the handle). The \lm\ then performs multi-viewpoint selection to identify the most optimal viewpoint free from occlusion. Finally, using the bounding box detection of the task-specific part of the object from the optimal viewpoint selected by the \lm\, we filter out a list of proposed candidate grasp poses and select the highest confidence grasp pose. This approach allows us to obtain the most optimal task-specific grasp pose, placement pose, and pre-action pose for non-prehensile tasks, all leveraging the capabilities of the VLM. After the action is generated, a simple motion planner can be used to move the robot to the desired pose as shown in detailed in Fig. \ref{fig:2}.

% For actions such as \emph{pick} and \emph{place}, we simply use the sampled graph pose as the waypoint and use a motion planner to guide the robot. For other actions, the \lm\ to writes code to synthesize the action. For example, the code for \emph{pull} can be intuitively defined as a retraction along the axis from which the grasp was approached. For action synthesis via code, we use the same multi-view reasoning. 

% For each action synthesized through code, we provide five example queries and corresponding responses as part of the prompt for few-shot prompting \cite{brown2020language} into the VLM for generating the code snippets.

% \begin{figure}[t]

%   \centering
%   \vspace{-1em}
%   \includegraphics[width=\linewidth]{figures/action_gen.png}
%   % Replace with your image file
%   \caption{\textbf{Action Generation Module.} \methodLong\ enables generation of two types of actions: object-centric and agent centric. For object-centric actions which require manipulation of an object, we leverage a foundation grasp model to generate all suitable grasps. Next, we leverage a VLM to detect the object from mult-view frames, and along with the candidate grasp poses and target subgoal, query the \lm\ to select the best view point. 
%   We filter and select the optimal grasp for the sub-task. For more agent-centric actions, the view-point selection process is the same, and the goal is to output code representing the change in pose of the end-effector from the current frame.}
%   \label{action_gen}

% \end{figure}

% % \vspace{-1em}

\subsection{Sub-task verification}
\label{sec:verifier}
\vspace{-1em}

To ensure that each sub-task $T_i$ is executed correctly, we introduce a \lm-based verifier. After every action for each sub-task are executed, we use the \lm\ to check if the end state matches the verifier condition $v_i$.
Similar to the action generation module, we use \textbf{multi-viewpoint \lm\ selection} to find the optimal view, avoiding errors due to occlusion or ambiguity from a single viewpoint. 
% We render $m$ views and leverage the \lm\ to select the optimal view for verification. 
If the verifier identifies failure, we re-attempt the action generation step for the previous sub-task from the current state.
Otherwise, the next sub-task $T_{i+1}$ is attempted. More details of the implementation is in the Appendix. 
% Once the final sub-task is verified, we manually measure whether the original task $T$ is completed.

% re-iterating the generation of primitive action code, or re-selecting viewpoints for action generation.

% This process also enables building of a repository of skills, where when the execution is successful, the skills will be appended to the library of skills. A skill can be re-used in future scenarios,

\vspace{-1em}

\section{Experiments}
\label{sec:experiments}
\vspace{-1em}

% \ranjay{Spread out the tables so that there is 1 on the top of each page.}
Our experiments are designed to address two questions: 1) Can \methodLong\ accurately solve a diverse set of tasks in a zero-shot manner? 2) Can data generated from \methodLong\ be used to train a robust policy?

\noindent \textbf{Implementation details.} 
We use both GPT-4V and Qwen-VL~\cite{bai2023qwen} as our \lm. We use GPT-4V for task decomposition, action generation, and verification. We use Qwen-VL to detect and extract object information. To ensure zero-shot execution within a reasonable budget, we limit the number of action steps in each trajectory to $50$ and the verification module allows a maximum of $30$ tries to accomplish a sub-goal.
For the task plan generation, we follow the prompting structure adapted from ProgPrompt \cite{singh2023progprompt}. All prompts input into the \lm\ are accompanied by few-shot demonstrations \cite{brown2020language}. Additionally, we provide three manually curated primitive action code snippets as examples to prompt the \lm\ for new action code generation. Full prompts are included in the Appendix. 
We use four viewpoints \( \mathbf{M}_4 = [front,wrist, left\_shoulder, right\_shoulder]\) for the simulation experiments, and re-render three viewpoints for the real-world experiments \cite{goyal2023rvt}. For better reasoning by the \lm, we use a resolution of \(256 \times 256\).

% For action generation, we provide a template prompt to guide the Vision-Language Model (VLM) in selecting the most optimal viewpoint for a given action. Additionally, a single image is created by concatenating the four viewpoints, each labeled with a number in sequence at the top left corner. For sub-task verification, we input the current verification condition \(v_i\) along with the concatenated image to prompt GPT-4V to determine the optimal viewpoint selection. 

%================================================
\vspace{-1em}
\subsection{Zero-shot Performance in Simulation}
\vspace{-1em}

We empirically study the zero-shot capability of \methodLong\ in solving 14 diverse tasks in simulation, covering a wide range of task configurations and action primitives for both prehensile and non-prehensile tasks. Our simulation experiments are reported to ensure reproducibility and provide a benchmark for future methods.

% TABLE ZERO-SHOT:
\begin{table}[t]
\centering
\caption{\textbf{Task-averaged success rate \% for zero-shot evaluation.} \methodLong\ outperformed other baselines in 10 out of 14 simulation tasks from RLBench \cite{james2020rlbench}. Each task was evaluated over 3 seeds to obtain the task-averaged success rate and standard deviations.}
\label{tab:zero-shot}
\begin{minipage}{\textwidth}
\centering
\renewcommand{\arraystretch}{1.2}
\setlength{\tabcolsep}{1.5pt}
\footnotesize
\begin{tabular}{@{}lccccccc@{}}
\toprule
Method & \scriptsize \texttt{Put\_block}  & \scriptsize \texttt{Play\_jenga} & \scriptsize \texttt{Open\_jar} & \scriptsize \texttt{Close\_box} & \scriptsize \texttt{Open\_box} & \scriptsize \texttt{Pickup\_cup} &\scriptsize 
\texttt{Push\_block}  \\
\midrule
VoxPoser \cite{huang2023voxposer} & 70.70$\pm 2.31$ & 0.00$\pm0.00$ & 0.00$\pm0.00$ & 0.00$\pm0.00$ & 0.00$\pm0.00$ & 26.70$\pm 14.00$ & \textbf{25.33$\pm 8.33$} \\
CAP \cite{liang2023code} & 84.00$\pm 16.00$ & 0.00$\pm0.00$ & 0.00$\pm0.00$ & 0.00$\pm0.00$ & 0.00$\pm0.00$ & 14.67$\pm 4.62$ & 8.00$\pm4.00$  \\
Scaling-up \cite{ha2023scaling} & 77.33$\pm6.11$& 0.00$\pm0.00$& 78.67$\pm11.55$ & 0.00$\pm0.00$ & 0.00$\pm0.00$ & 9.33$\pm2.26$ & 5.33$\pm6.11$ \\
MA (Ours) & \textbf{96.00$\pm 4.00$} & \textbf{77.33$\pm 6.11$} & \textbf{80.00$\pm 4.00$} & \textbf{33.33$\pm 12.86$} & \textbf{29.00$\pm 10.07$} & \textbf{82.67$\pm 14.04$} & 20.00$\pm 4.00$  \\
\bottomrule

\toprule
\footnotesize

Method & \scriptsize

\texttt{Take\_umbrella} & \scriptsize

\texttt{Sort\_mustard} & \scriptsize

\texttt{Open\_wine} & \scriptsize

\texttt{Lamp\_on} &\scriptsize

 \texttt{Put\_knife} &\scriptsize

\texttt{Pick\_\&\_lift} & \scriptsize

\texttt{Insert\_block}\\
\midrule
\footnotesize

VoxPoser\cite{huang2023voxposer} & 33.33$\pm 8.33$ & \textbf{96.00$\pm 6.93$} & 8.00$\pm4.00$ & 57.30$\pm12.22$ & \textbf{92.00$\pm4.00$} & 96.00$\pm0.00$ & 0.00$\pm0.00$\\
\footnotesize

CAP\cite{liang2023code} & 4.00$\pm4.00$ & 0.00$\pm0.00$ & 0.00$\pm0.00$ & 64.00$\pm6.93$ & 14.67$\pm8.33$ & \textbf{100.00$\pm 0.00$} & 0.00$\pm0.00$\\
\footnotesize

Scaling-up \cite{ha2023scaling} & 6.67$\pm2.31$& 41.33$\pm12.86$ & 33.33$\pm20.13$& 60.00$\pm8.00$ & 24.00$\pm0.00$ & \textbf{100.00$\pm0.00$} & 0.00$\pm0.00$ \\
MA (Ours) & \textbf{61.33$\pm20.13$} & 64.00$\pm6.93$ & \textbf{42.00$\pm4.00$} & \textbf{69.33$\pm6.11$} & 52.00$\pm10.58$ & 84.00$\pm6.93$ & \textbf{33.33$\pm4.62$}\\
\bottomrule
\end{tabular}
\end{minipage}
\vspace{-1em}

\end{table}

\noindent \textbf{Environment and tasks.} The simulation is set up in CoppeliaSim and interfaced through PyRep. All simulation experiments use a Franka Panda robot with a parallel gripper. Input observations are captured from four RGB-D cameras positioned around a tabletop setting. 
We use RLBench~\cite{james2020rlbench}, a robot learning benchmark with diverse tasks conditioned on language and provided success conditions. 
We sample $14$ tasks from RLBench, covering a diverse range of action primitives, task horizons, and object position perturbations. 
Each action can be represented as a way-point, and the trajectories are computed and executed via a motion planner using the Open Motion Planning Library \cite{sucan2012open}.

\textbf{Baselines.} We compare against three state-of-the-art zero-shot data generation approaches: Code-as-Policies (CAP)~\cite{liang2023code}, Scaling-up-Distilling-Down (Scaling-up) \cite{ha2023scaling} and VoxPoser~\cite{huang2023voxposer}. CAP uses language models to generate executable programs that call hand-crafted primitive actions. 
VoxPoser~\cite{huang2023voxposer} builds a 3D voxel map of value functions for predicting waypoints. Scaling-up \cite{ha2023scaling} is a Large Language Model (LLM) equipped with a suite of 6 DoF exploration primitives, which it uses within its framework to generate robotic data for distilling a policy.
We provided CAP with ground truth simulation state information and object models, and we supplied VoxPoser with ground truth segmented object point clouds from the simulator. This inherently puts \methodLong\ at a disadvantage in comparison.
% We used the same set of language instructions for all evaluated methods corresponding to the given tasks.

\noindent\textbf{Results: }
\textbf{\methodLong\ can generate successful trajectories for all $14$ tasks while Scaling-up, VoxPoser and CAP cover only $10$,  $9$ and $7$} tasks, respectively (Table~\ref{tab:zero-shot}). Without the privileged state information, the baselines would not succeed on any of the $14$ tasks.
\methodLong\ outperforms the baselines in $10$ out of the $14$ tasks. The three tasks where our method achieves lower performance require fine-grained manipulation of objects, which are the hardest task without the privileged state information used by baselines. 
VoxPoser fails in the tasks that require moving the arm beyond 4-DoF. 
\methodLong\ outperforms the strongest baseline, VoxPoser, by an average task-averaged margin of up to $22\%$.

% \begin{wrapfigure}{r}{0.3\textwidth}
%   \centering
%   \includegraphics[width=\linewidth]{figures/scaling_law.png}
%   \caption{We demonstrate that the data generated using \methodLong\ enhances model performance when scaled with an increased number of demonstrations. Notably, the quality of our generated data exhibits a better rate of improvement in performance compared to human-generated data as the scale of demonstrations increases.}
%   \label{fig:scaling}
% \end{wrapfigure}

\vspace{-1em}
%================================================
\subsection{Behavior cloning with demonstrations from \methodLong}
\vspace{-1em}

Next, we analyze the quality of the generated data by comparing the success rates of behavior cloning models trained with the data. Zero-shot methods like \methodLong\ are computationally expensive but hold the potential to generate useful training data. To evaluate the quality and effectiveness of the generated training data, we use the methods described in the previous section to generate data for each task. We also compare performance against a model trained on human-generated demonstrations across the 12 tasks. We use the data to train behavior cloning policies.

% TABLE: QUALITATIVE
\begin{table}[t]
\centering
\caption{\textbf{Behavior Cloning with different generated data.} The behavior cloning policy trained on the data generated by \methodLong\ provides the best performance on 10 out of 12 tasks compared to the other autonomous data generation baselines. We report the Success Rate \% for behaviour cloning policies trained with data generated from VoxPoser \cite{huang2023voxposer} and Code as Policies \cite{liang2023code} in comparison. Note that the RLBench\cite{james2020rlbench} baseline uses human expert demonstrations and is considered an upper bound for behavior cloning.
% \ranjay{For all figures and tables: the first sentence should be the main take away message. The next sentences can describe the table. Update all the figure and table captions please. Also bold the numbers that are best.}
}
\label{tab:qualitative}
\begin{minipage}{\textwidth}
\centering
\renewcommand{\arraystretch}{1.2}
\setlength{\tabcolsep}{1.5pt}
\footnotesize
\begin{tabular}{@{}lccccccc@{}}
\toprule
\footnotesize

Data &\footnotesize
 Models &\footnotesize
 \texttt{Put\_block} & \footnotesize
\texttt{Play\_jenga} &\footnotesize
 \texttt{Open\_jar} &\footnotesize
 \texttt{Close\_box} &\footnotesize
 \texttt{Open\_box} & \footnotesize
\texttt{Pickup\_cup} \\
\midrule
\footnotesize
VoxPoser\cite{huang2023voxposer} & PerAct\cite{shridhar2023perceiver} & \scriptsize 2.67$\pm 2.31$  & - &  - & -  & -  & \scriptsize 4.00$\pm 4.00$ \\
CAP\cite{liang2023code} &PerAct\cite{shridhar2023perceiver} &\scriptsize 6.67$\pm 2.31$  & - &  - & -  & -  & \scriptsize 14.67$\pm 12.86$ \\
Scaling-up \cite{ha2023scaling} &PerAct\cite{shridhar2023perceiver} &\scriptsize 22.67$\pm 15.14$   & - & \scriptsize5.33$\pm 6.11$  & -  & -  & \scriptsize  14.67$\pm 2.31$  \\
MA (Ours) & PerAct\cite{shridhar2023perceiver} & \scriptsize \textbf{85.33$\pm 10.07$}  & \scriptsize 81.33$\pm 2.31$  & \scriptsize 21.33$\pm 10.07$ & \scriptsize 42.67$\pm 8.33$ &  \scriptsize \textbf{30.67$\pm 11.55$} & \scriptsize 54.00$\pm 12.49$\\ 
RLBench\cite{james2020rlbench} & PerAct\cite{shridhar2023perceiver} & \scriptsize 20.00$\pm 18.33$ & \scriptsize \textbf{81.33$\pm 9.24$} & \scriptsize \textbf{58.67$\pm 45.49$} & \scriptsize \textbf{68.00$\pm 24.98$} & \scriptsize 14.67$\pm 6.11$ & \scriptsize \textbf{54.67$\pm 23.09$} \\
\hline
VoxPoser\cite{huang2023voxposer} & RVT-2\cite{goyal2024rvt} & \scriptsize 73.33$\pm 2.31$  & - &  - & -  & -  & \scriptsize 2.67$\pm 2.31$ \\
CAP\cite{liang2023code} & RVT-2\cite{goyal2024rvt}  &\scriptsize 78.66$\pm 8.32$  & - &  - & -  & -  & \scriptsize 77.33$\pm 19.73$ \\
Scaling-up \cite{ha2023scaling} &RVT-2\cite{goyal2024rvt} &\scriptsize 38.67$\pm 2.31$  & - &  \scriptsize33.33$\pm 2.31$ & -  & -  &\scriptsize92.00$\pm 4.00$ \\
MA (Ours) & RVT-2\cite{goyal2024rvt} & \scriptsize 85.33$\pm 2.31$  & \scriptsize 82.67$\pm 2.31$  & \scriptsize 78.67$\pm 10.06$ & \scriptsize \textbf{82.67$\pm 2.31$} & \scriptsize \textbf{24.00$\pm 4.00$} & \scriptsize 97.33$\pm 2.31$\\
RLBench\cite{james2020rlbench} & RVT-2\cite{goyal2024rvt} & \scriptsize \textbf{86.67$\pm 2.31$} & \scriptsize \textbf{85.33$\pm 2.31$} & \scriptsize \textbf{81.33$\pm 6.11$} & \scriptsize 76.00$\pm 4.00$ & \scriptsize 4.00$\pm 4.00$ & \scriptsize \textbf{97.33$\pm 2.31$} \\
\bottomrule

\toprule
\footnotesize

Data & Models &
\footnotesize
\texttt{Take\_umbrella} & 
\footnotesize
\texttt{Sort\_mustard} & 
\footnotesize
\texttt{Open\_wine} &\footnotesize
 \texttt{Lamp\_on} & \footnotesize
\texttt{Put\_knife} & \footnotesize
\texttt{Pick\_\&\_lift} \\
\midrule
\footnotesize
VoxPoser\cite{huang2023voxposer} & PerAct\cite{shridhar2023perceiver} & \scriptsize 4.00$\pm 4.00$ & \scriptsize 0.00$\pm 0.00$ & \scriptsize 1.33$\pm 2.31$  & \scriptsize 5.33$\pm 4.62$  & \scriptsize 1.33$\pm 2.31$  & \scriptsize 5.67$\pm 1.64$  \\
CAP\cite{liang2023code} & PerAct\cite{shridhar2023perceiver} & \scriptsize 13.33$\pm 10.06$  & - &  - & \scriptsize 8.00$\pm 16.00$  & \scriptsize 9.33$\pm 6.11$  & \scriptsize 46.67$\pm 2.31$ \\
Scaling-up \cite{ha2023scaling} &PerAct\cite{shridhar2023perceiver} &\scriptsize 4.00$\pm 4.00$  & \scriptsize0.00$\pm 0.00$  &  \scriptsize81.33$\pm 12.86$  & \scriptsize76.00$\pm 4.00$  & \scriptsize5.33$\pm 2.31$ & \scriptsize 53.33$\pm 10.06$  \\
MA (Ours) & PerAct\cite{shridhar2023perceiver} & \scriptsize \textbf{84.00$\pm 6.93$} & \scriptsize 53.33$\pm 6.11$ & \scriptsize 86.67$\pm 6.11$ & \scriptsize \textbf{89.33$\pm 6.11$} & \scriptsize 8.00$\pm 4.00$ & \scriptsize 33.33$\pm 2.31$ \\ 
RLBench\cite{james2020rlbench} & PerAct\cite{shridhar2023perceiver} & \scriptsize {58.67$\pm 50.80$} & \scriptsize \textbf{53.33$\pm 34.02$} & \scriptsize \textbf{86.67$\pm 12.86$} & \scriptsize 84.00$\pm 13.86$ &  \scriptsize \textbf{30.67$\pm 10.07$}& \scriptsize \textbf{62.67$\pm 9.24$} \\
 \hline
 VoxPoser\cite{huang2023voxposer} & RVT-2\cite{goyal2024rvt} & \scriptsize 5.33$\pm 6.11$ & \scriptsize 1.33$\pm 2.31$ & \scriptsize 1.33$\pm 2.31$  & \scriptsize 2.67$\pm 2.31$  & \scriptsize 1.33$\pm 2.31$  & \scriptsize 17.33$\pm 2.31$  \\
 CAP\cite{liang2023code} & RVT-2\cite{goyal2024rvt} & \scriptsize 89.33$\pm 6.11$  & - &  - & \scriptsize 85.33$\pm 8.32$  & \scriptsize 52.00$\pm 10.58$  & \scriptsize 82.66$\pm 20.53$ \\
 Scaling-up \cite{ha2023scaling} &RVT-2\cite{goyal2024rvt}    & \scriptsize 94.67$\pm 4.62$     &\scriptsize24.00$\pm 4.00$   &\scriptsize62.67$\pm 2.31$   & \scriptsize21.33$\pm 2.31$& \scriptsize53.33$\pm 2.31$ &\scriptsize80.00$\pm 6.93$ \\
MA (Ours) & RVT-2\cite{goyal2024rvt} & \scriptsize 94.67$\pm 2.31$ & \scriptsize \textbf{73.33$\pm 2.31$} & \textbf{\scriptsize 93.33$\pm 6.11$} & \scriptsize 84.00$\pm 10.58$ & \scriptsize 69.33$\pm 12.85$  & \scriptsize \textbf{82.67$\pm 12.22$} \\
RLBench\cite{james2020rlbench} & RVT-2\cite{goyal2024rvt} & \scriptsize \textbf{97.33$\pm 2.31$} & \scriptsize 69.33$\pm 8.33$  & \scriptsize 88.00$\pm 8.00$ & \scriptsize \textbf{93.33$\pm 4.62$} & \scriptsize \textbf{72.00$\pm 10.58$} & \scriptsize 64.00$\pm 10.58$ \\
\bottomrule
\vspace{-3em}

\end{tabular}
\end{minipage}
\end{table}
% \vspace{-1em}

\noindent \textbf{Data generation details.} We generate 10 successful demonstrations per task. We use the system's success condition to filter for successful demonstrations. 
Each of the demonstrations consist of a language instruction, RGB-D frames for the trajectory, and waypoints represented as 6 DoF gripper poses and states. For the tasks that the baselines were unable to generate any successful demonstrations, we patched the missing training data with RLBench system-generated demonstrations. 

\noindent \textbf{Training and evaluation protocol.} We train two models using the generated demonstrations: the Perceiver-Actor (PerAct) model~\cite{shridhar2023perceiver}, a transformer-based robotic manipulation behavior cloning model that expects tokenized voxel grids and language instructions as inputs and predicts discretized voxel grid 6 DoF poses and gripper states, and also RVT-2 model~\cite{goyal2024rvt}, a multi-view transformer-based BC model. The RVT-2 model uses tokenized image patches and CLIP-encoded language instructions as input to predict keypoint actions as translation heatmaps, discretized rotation in Euler angles, and the gripper's binary state. Notably, RVT-2 is currently the highest-performing model on the RLBench benchmark.
For all the generated training datasets, we train a multi-task PerAct policy with a batch size of $4$ for $30$k iterations on a single RTX A100, and RVT-2 with a batch size of 24 for $3.3$k iterations on 4 A100s. To ensure consistent evaluation, we generate one set of testing environments with RLBench. We evaluate the last checkpoint from each of the trained policies. Each policy is evaluated for $25$ episodes across each task using $3$ different seeds. We measure the success rate based on the simulation-defined success condition.

\begin{figure}[t]
  \centering
  \includegraphics[width=0.9\linewidth]{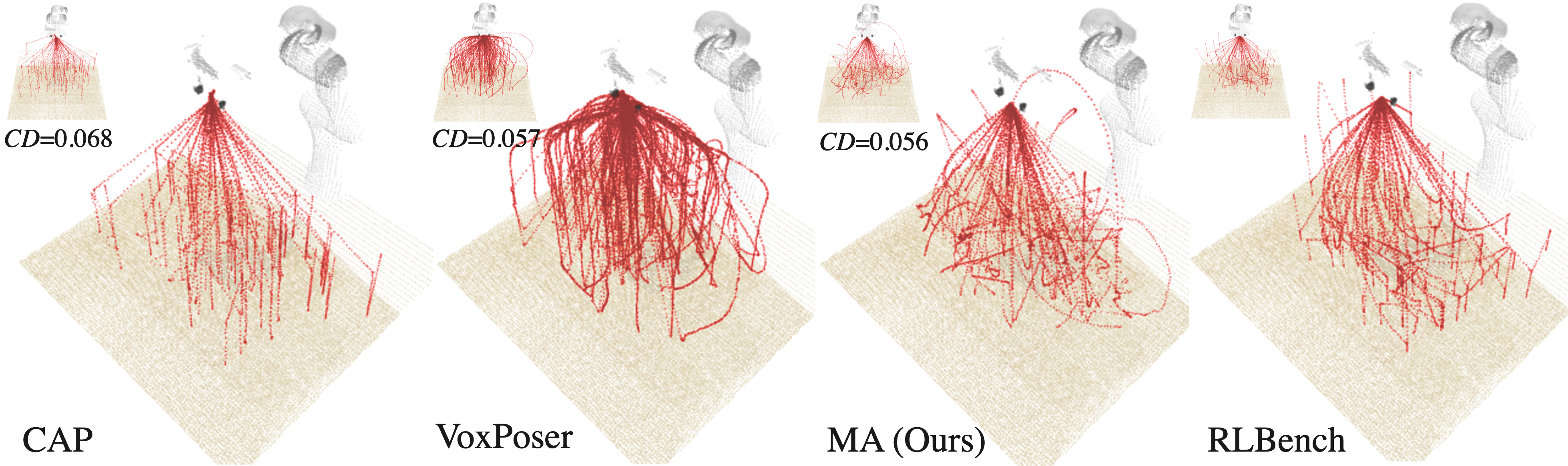}
  % Replace with your image file
  \caption{\textbf{Action Distribution for Generated Data:} We compare the action distribution of data generated by various methods against human-generated demonstrations via RLBench on the same set of tasks. We observed a high similarity between the distribution of our generated data and the human-generated data. This is further supported by the computed CD between our methods and the RLBench data, which yields the lowest (CD=0.056).}
  \label{fig:distribution}
  \vspace{-1.5em}
\end{figure}

\noindent \textbf{Results: } 
\textbf{Policies trained using \methodLong\ data perform similarly to policies trained using hand-scripted demonstrations ($p=0.973$) for PerAct (Table~\ref{tab:qualitative})}.
Training on either \methodLong\ or on hand-scripted demonstrations results in a performance difference of a mere $0.27$\% across all tasks. 
Furthermore, models trained on data from the baselines exhibit a statistically lower performance ($p\le 0.01$ for VoxPoser, Scaling-up and CAP). 
One of the main factors potentially contributing to the differences in the performance could be that \methodLong\ generates diverse expert trajectories that are preferable to humans. This can be seen in Fig. \ref{fig:distribution}, which shows the action distribution of the generated data by different methods for the same given tasks. Additionally, our generated data recorded the lowest Chamfer Distance (CD) of 0.056 with human-generated demonstrations data.
We also observed that the policy trained on MA data achieves a lower standard deviation of $3.39$ across all tasks, compared to the zero-shot performance standard deviation of $8.48$. This suggests the benefits of training over generated data instead of relying solely on zero-shot deployment.

\begin{wrapfigure}{r}{0.4\textwidth}
%   \vspace{-1.3em}
%   \vspace{-2.3em}
%   \vspace{-4.3em}
  
  \begin{center}
    % \hspace{2.5cm}
    \includegraphics[width=0.4\textwidth]{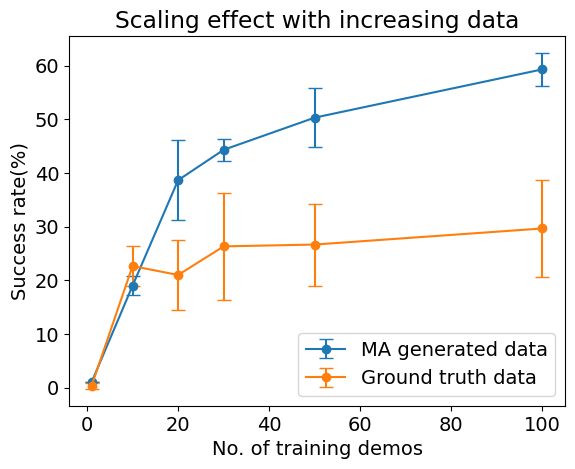}

%   \vspace{-1.2em}
  \caption{\textbf{Scaling experiment.} Scaling effect of model performance with increasing training demonstrations.} 
  \label{fig:ablations}
  \end{center}
%   \vspace{-1.7em}
%   \vspace{-0.5em}
\end{wrapfigure}

% \begin{wrapfigure}{r}{0.3\textwidth}
% \vspace{-5em}
%   \centering
%   \includegraphics[width=\linewidth]{figures/scaling_law.png}
%   \caption{Scaling Ablation.}
%   \label{fig:scaling}
%   \vspace{-2em}
% \end{wrapfigure}

\vspace{-1em}

%================================================
\subsection{Real-world experiments}
\vspace{-1em}

Finally, we evaluate \methodLong\ in the real world.
We also automatically generate real-world demonstrations for training PerAct. 

\textbf{Environment and tasks.} We employ a Franka Panda manipulator equipped with a parallel gripper. We use a front-facing Kinect 2 RGB-D camera. To generate multi-view inputs for the \methodLong\ framework, we re-render virtual viewpoints from the generated point cloud, similar to prior work~\cite{goyal2023rvt}. 
We selected 7 representative real-world tasks, both prehensile and non-prehensile: \texttt{open\_jar}, \texttt{sort\_objects}, \texttt{correct\_dices}, \texttt{open\_drawer}, \texttt{on\_lamp}, \texttt{press\_switch}, and \texttt{close\_laptop}, all conditioned on language instructions. Each task was evaluated over 10 episodes with varying object poses across 3 trials.

\noindent \textbf{Data generation details.} We used \methodLong\ to generate $6$ demonstrations for each task and manually perform scene resets when failures occur. We train a similar multi-task PerAct for $120$k iterations and evaluate the trained policies in a manner similar to the zero-shot experiments.

% TABLE: REAL-WORLD
\begin{table}[t]
\centering
\scriptsize
\setlength{\tabcolsep}{1.5pt}

\caption{\textbf{Real-world Results.} The model trained on the data generated by our model in the real world (no expert in the loop) demonstrates on par results with the model trained on human expert collected data. We present a comparison of success rates for task completion in a zero-shot manner (Code as Policies \cite{liang2023code} and \methodLong), and using trained policies from \methodLong\ data and human expert data.}
\begin{tabular}{cccccccc}
\toprule
    & Open\_drawer & Sort\_object & On\_lamp & Open\_jar & Correct\_dice & Press\_switch & Close\_laptop\\
    \midrule
    CAP (0-shot) & $0.00\pm0.00$ &  $13.33\pm5.77$  &  $0.00\pm0.00$  & $6.67\pm5.77$  & $6.67\pm5.77$ & $0.00\pm0.00$ & $0.00\pm0.00$ \\
    MA (0-shot) & \textbf{36.67$\pm5.77$} & \textbf{60.00$\pm10.00$} & \textbf{26.67$\pm11.55$} &\textbf{40.00$\pm10.00$} & \textbf{53.33$\pm5.77$} & \textbf{20.00$\pm10.00$} & \textbf{33.33$\pm5.77$}\\ \midrule
    PerAct { (MA data)} & $50.00\pm0.00$ & $33.33\pm5.77$ & $50.00\pm0.00$ &$ 56.67\pm5.77$ & $ 60.00\pm0.00$ & \textbf{$ 56.67\pm5.77$} & $33.33\pm5.77$\\
    PerAct {(Human data)} & \textbf{53.33$\pm11.55$} & \textbf{36.67$\pm5.77$} & \textbf{60.00$\pm0.00$} &\textbf{76.67$\pm5.77$} & \textbf{80.00$\pm10.00$} & 33.33$\pm5.77$ & \textbf{53.33$\pm5.77$}\\
\bottomrule
\end{tabular}
\vspace{-1em}
\label{tab:real-world}
\end{table}

\noindent \textbf{Results:} \textbf{\methodLong\ is able to generate successful demonstrations for each of the $7$ real world tasks.}
Even for the worst-performing task, \methodLong\ achieves a success rate of more than $25\%$. Our approach outperforms \textbf{CAP} by 38\%.  Consistent with the simulation results, \textbf{training with the data generated by \methodLong\ produces a more robust policy} compared to performing zero-shot. Additionally, in 4 out of 5 tasks, the trained policies perform better than the zero-shot approach. 
The policy under-performs on the \texttt{sort\_object} task, because it requires longer-horizon memory
% has to remember which objects have already been sorted
—a known limitation pointed out in PerAct~\cite{shridhar2023perceiver}.

\begin{wrapfigure}{r}{0.4\textwidth}
%   \vspace{-1.3em}
%   \vspace{-2.3em}
%   \vspace{-4.3em}
  
  \begin{center}
    % \hspace{2.5cm}
    \includegraphics[width=0.4\textwidth]{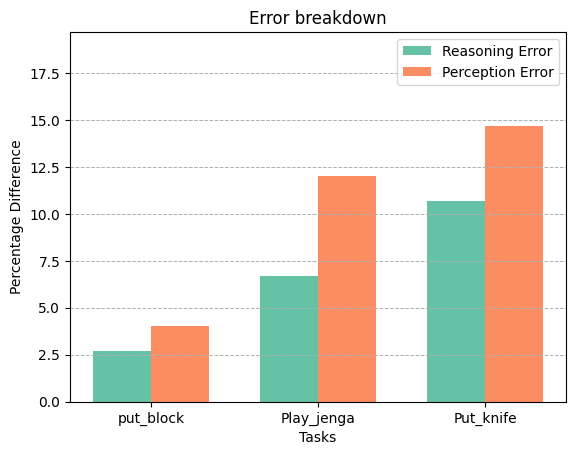}

%   \vspace{-1.2em}
  \caption{\textbf{Error breakdown.} Error breakdown for three tasks from simulation.} 
  \label{fig:ablations}
  \end{center}
  \vspace{-2em}
%   \vspace{-1.7em}
%   \vspace{-0.5em}
\end{wrapfigure}

% \begin{figure}[t]
%   \centering
%   \includegraphics[width=.4\linewidth]{figures/scaling_law.png}
%   % Replace with your image file
%   \caption{Scaling.}
%   \label{fig:scaling}
% \end{figure}

% \begin{wrapfigure}{r}{0.3\textwidth}
% \vspace{-5em}
%   \centering
%   \includegraphics[width=\linewidth]{figures/scaling_law.png}
%   \caption{We demonstrate that the data generated using \methodLong\ significantly enhances model performance when scaled with an increased number of demonstrations. Notably, the quality of our generated data exhibits a better rate of improvement in performance compared to human-generated data as the scale of demonstrations increases.}
%   \label{fig:scaling}
%   \vspace{-2em}
% \end{wrapfigure}

\vspace{-1em}

\subsection{Ablations}
\vspace{-1em}
For effective real-world deployment of \methodLong,\ it's crucial that the collected data supports scaling of robotics transformers and offers diverse skills and interacted objects. We conducted an ablation study to evaluate the quality of \methodLong-generated data for scaling and its generalization to language instruction changes. For scaling, we generated behavior cloning data, ranging from 1 to 100 training demonstrations from RLBench and \methodLong\ for a single task (\texttt{put\_block}), and trained a PerAct policy. For generalization, we varied the \texttt{sort\_mustard} task with different language instructions and target objects. We compared our approach to VoxPoser to assess robustness to object and language instruction changes. Further implementation details are in the supplementary materials. \textbf{Result:} Our scaling experiments demonstrate that generating more training data via \methodLong\ improves PerAct policy performance (Fig.~\ref{fig:ablations}). The data from our approach shows a better rate of change with a slope of 0.503 for a linear fit, compared to 0.197 for RLBench-generated data. Additionally, \methodLong\ data is more generalizable and robust to language instruction changes, outperforming VoxPoser in task success across language and object variations. Detailed results in the appendix.

\subsection{Error breakdown}
We examine the potential errors introduced by each respective VLM and explore ways to further improve the overall system. We conducted simulation experiments where tasks could be easily reset, focusing on two major components where VLMs make decisions: "Perception error" and "Reasoning error." "Perception error" refers to mistakes made by the VLM in detecting target objects and selecting optimal viewpoints for generating task-specific grasp pose, primarily related to the selection of the Multi-viewpoint VLM. "Reasoning error" involves the Sub-task Verification Module, where the VLM is responsible for deciding if sub-tasks have been successfully completed. Due to resource constraints, we conducted experiments on the \texttt{play\_jenga} task. We systematically replaced the VLMs in each component with a human, allowing the human to make decisions. By comparing the system's performance with human decision-making to that with VLMs, we were able to quantify the errors caused by the VLMs.

% \ranjay{Avoid the term ``scaling law''.}

% \subsection{Scaling effect with generated data}
% In contrast to manually collected robot demonstrations \cite{open_x_embodiment_rt_x_2023}, our objective is to demonstrate that \methodLong\ can be utilized as a large-scale data generator with the associated cost of employing proprietary VLMs. We hypothesize that the data generated by this method adheres to the scaling laws of training robotic transformers. To evaluate this, we conducted an analytical study on a specific simulation task, \texttt{put\_block}. We generated both ground truth data from RLBench and data using \methodLong\ at scales of [1, 10, 20, 30, 50, 100]. For each dataset, we trained a PerAct policy for same epoch and evaluated its performance across 100 episodes of a testing dataset.

\vspace{-1em}
\section{Discussion}
\vspace{-1em}
\textbf{Limitations.}
Despite the promising results, \methodLong\ has several limitations. First, \methodLong\ depends on the availability of large language models (LLMs), which introduces reliance on foundational models. However, with the ongoing development of open-source LLMs, this issue is likely to diminish over time. Second, \methodLong\ struggles with dynamic manipulation tasks and non-prehensile tasks, as it cannot generate alternative grasp poses for these scenarios. Third, the system's highly modular nature, integrating multiple VLMs, can lead to compounding errors when generating zero-shot trajectories. Emerging specialized VLMs \cite{yuan2024robopoint} and others which may help address this issue. Finally, manual prompt engineering for in-context learning is still required. Nonetheless, recent advancements in alignment and prompting techniques \cite{ouyang2022training,wang2022self, wei2022chain} offer potential solutions to reduce the effort involved in prompt engineering. 

\textbf{Conclusion.} \methodLong\ is a scalable environment-agnostic approach for generating zero-shot demonstration for robotic tasks without the use of privileged environment information. \methodLong\ uses LLMs to do high level planning and scene understanding and is capable of error recovery. This enables high quality data generation for behavior cloning that can achieve better performance that using human data.

\bibliography{main}

\end{document}